\documentclass[journal,twoside,web]{ieeecolor}

\usepackage{etoolbox}
\makeatletter
\@ifundefined{color@begingroup}%
{\let\color@begingroup\relax
\let\color@endgroup\relax}{}%
\def\fix@ieeecolor@hbox#1{%
\hbox{\color@begingroup#1\color@endgroup}}
\patchcmd\@makecaption{\hbox}{\fix@ieeecolor@hbox}{}{\FAILED}
\patchcmd\@makecaption{\hbox}{\fix@ieeecolor@hbox}{}{\FAILED}

\usepackage{generic}
\usepackage{cite}
\usepackage{amsmath,amssymb,amsfonts}
\usepackage{graphicx}
\usepackage{algorithm,algorithmic}
\usepackage{hyperref}
\hypersetup{hidelinks=true}
\usepackage{textcomp}

\usepackage{longtable}
\usepackage{array}
\usepackage{multirow}
\usepackage{colortbl}
\usepackage{xspace}
\usepackage{arydshln}
\usepackage{float}
\usepackage{makecell}
\usepackage{bbding}
\usepackage{xcolor}
\usepackage{soul}
\usepackage{empheq}
\usepackage{paralist}
\usepackage{tabularx}

\def \ie {\emph{i.e.}~}
\def \eg {\emph{e.g.}~}
\def \etc {\emph{etc.}~}
\def \etal {\emph{et al.}~}

\newcommand{\myparagraph}[1]{\noindent\textbf{#1.}~}

\def\BibTeX{{\rm B\kern-.05em{\sc i\kern-.025em b}\kern-.08em
    T\kern-.1667em\lower.7ex\hbox{E}\kern-.125emX}}
\begin{document}
\title{Multimodal Large Language Model driven Radiology Report Generation with Clinical Knowledge Enhancement}
\author{Miaojing Shi, Tianyu Cen, Zijie Yue, Meng Wei, Oluwatosin Alabi and Tom Vercauteren
\thanks{M.Shi and Z.Yue are with the College of Electronic and Information Engineering, Tongji University, Shanghai, China (Corresponding
author: Z. Yue) (e-mail: mshi@tongji.edu.cn; zijie@tongji.edu.cn)}
\thanks{T.Cen is with Shanghai Research Institute for Intelligent Autonomous Systems, Tongji University, Shanghai, China (e-mail: chris\_cen@tongji.edu.cn)}
\thanks{M. Wei, O. Alabi, and T. Vercauteren are with the School of Biomedical Engineering \& Imaging Sciences, King’s College London, London, UK (e-mail: meng.wei@kcl.ac.uk; oluwatosin.alabi@kcl.ac.uk;
tom.vercauteren@kcl.ac.uk)}
\thanks{This work was supported by China Fundamental Research Funds
for the Central Universities, King’s Cambridge 1 Access Fund, National
Natural Science Foundation of China under Grant 62401393, and core
funding from the Wellcome/EPSRC [WT203148/Z/16/Z; NS/A000049/1].}
\thanks{T. Vercauteren is supported by a Medtronic / RAEng Research Chair [RCSRF1819\textbackslash7\textbackslash34], and is co-founder and shareholder of Hypervision Surgical. M. Wei and O. Alabi are supported by the EPSRC CDT [EP/S022104/1].}}

\maketitle

\begin{abstract}
Radiology report generation (RRG) has attracted significant attention due to its potential to reduce the workload of radiologists.
The performance of current RRG approaches remains unsatisfactory against clinical standards.
This paper introduces a novel RRG method, \textbf{MLLM-RRG}, that integrates multimodal large language models (MLLMs) with various types of clinical knowledge to generate accurate and comprehensive chest X-ray reports.
Our method first designs a referring anatomical feature extractor that leverages anatomical knowledge to analyze different regions of the chest X-ray image and extract visual features without explicitly detecting regions. 
Next, based on the MLLM's decoder, we develop a multimodal report generator that leverages multimodal prompts constructed from dedicated visual features and textual instructions to produce the radiology report in an auto-regressive way.
Finally, we introduce a disease-oriented clinical classification and alignment scheme in a multi-task learning manner to leverage disease knowledge to better preserve the clinical relevance among the generated reports.
Once the model is trained, we also introduce a novel clinical quality reinforcement learning strategy to enhance the MLLM with report knowledge, further refining the tones of the generated reports towards radiologists.
Extensive experiments on the MIMIC-CXR and IU X-Ray datasets demonstrate the superiority of our method over the state of the art.
Our codes will be available at https://github.com/viscom-tongji/MLLM-RRG.
\end{abstract}


\section{Introduction}
Chest X-ray radiology report generation (RRG) aims to automatically generate radiology reports based on given images, ensuring the accuracy and fluency of the reports.
Since this technology could significantly reduce the time that radiologists spend on writing reports, recent advances in multimodal models have renewed the attention on RRG from both the radiology and computer science communities
\cite{chen2020generating,qin2022reinforced,jin2024promptmrg,gu2024complex,tanida2023interactive,li2023unify,li2023dynamic}.
The core technique of radiology report generation stems from image captioning~\cite{stefanini2022show} in a general computer vision view.
Unlike image captioning, RRG not only focuses on reporting the disease-related symptoms across different anatomical regions but also requires reasoning from these symptoms.

\begin{figure}[!t]
\includegraphics[width=\columnwidth]{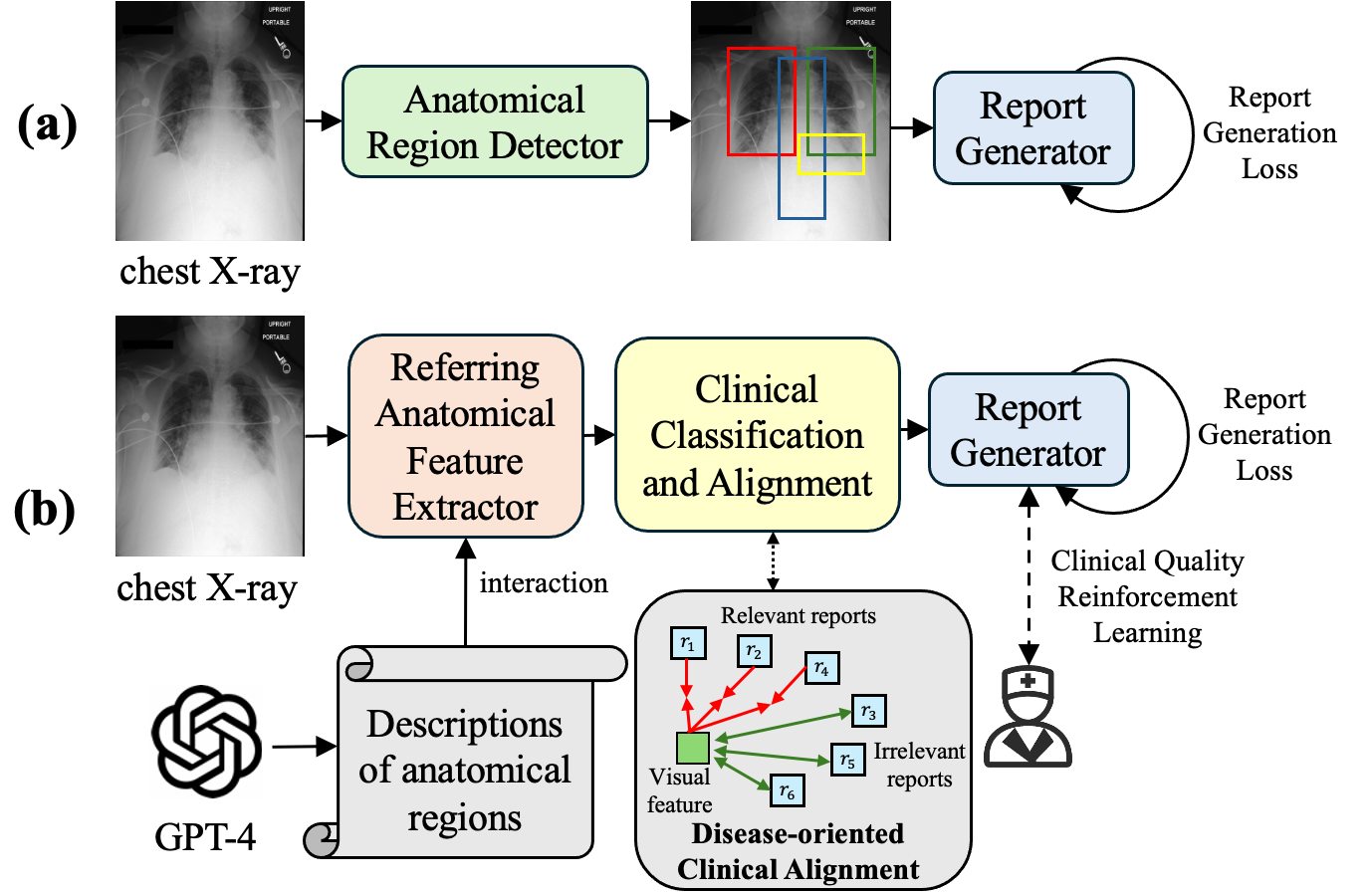}
\caption{Given a chest X-ray image, (a) previous RRG methods often rely on visual features alone to generate reports 
(b) In contrast, we offer new components such as referring anatomical feature extractor, clinical classification and alignment, clinical quality reinforcement learning to enhance the clinical knowledge of the model and consequently improve the clinical relevance of the generated reports.\vspace{-1.5em}}
\label{fig:openfig}
\end{figure}

However, existing methods in the RRG domain remain some distance from being trusted and readily deployed in clinical practice. One major reason lies in the fact that they rely solely on visual cues in chest X-ray images to generate reports, as shown in Figure~\ref{fig:openfig}(a), visual cues alone are often insufficient to support reliable clinical judgements. In contrast, expert radiologists compose reports based on both visual cues and their extensive clinical knowledge, the latter of which often plays a decisive role in producing accurate and well-structured reports. We categorize three types of clinical knowledge that radiologists use:

\myparagraph{\textit{Anatomical knowledge}}
Anatomical knowledge refers to the understanding of chest structures, including locations and boundaries of key anatomical regions (\eg, lungs, cardiac silhouette). Radiologists use anatomical knowledge to systematically locate and examine different anatomical regions when interpreting chest X-ray images.
To replicate this procedure, 
recent methods ~\cite{tanida2023interactive,gu2024complex} introduce pre-trained segmentation models or region detectors to extract features from specific regions in a chest X-ray image. 
However, these methods are constrained by the anatomical categories learned in pre-trained segmenters or detectors, limiting their flexibility to a wide coverage of anatomical regions. 
Furthermore, these pre-trained models prevent end-to-end training of the whole pipeline, whilst segmentation or detection errors can propagate and accumulate to subsequent modules, ultimately degrading report generation performance. 

\myparagraph{\textit{Disease knowledge}}
Disease knowledge refers to the understanding of different diseases in chest X-ray domain (\eg, pneumothorax typically appears as a visible pleural line with absence of lung markings peripheral to it). Through years of clinical accumulation, radiologists can associate the examined case with clinically similar cases encountered before, so that their prior diagnostic experience can be utilized to support the interpretation of the current case. 
Inspired by this, some RRG methods~\cite{tao2024memory, liu2024bootstrapping, lin2025dc} retrieve similar cases from history based on image-to-image or image-to-text similarities and incorporate the retrieved information into the representation learning or decoding stage. 
However, since the retrieval process is not guided by disease labels, the retrieved cases may not truly be disease-relevant. 
More recently, Park \etal~\cite{park2025dart} introduce a disease-matching constraint to ensure that retrieved cases exhibit exact same diseases as the examined case. 
Nevertheless, their matching constraint is rather strict. 
Clinically correlated cases may be mistakenly treated as irrelevant
hence valuable diagnostic information is missed from relevant cases.

\myparagraph{\textit{Report knowledge}}
Report knowledge governs how clinical findings should be documented in a professional radiology report. 
Radiologists use such knowledge to write reports that are both clinically accurate and professionally standardized, which is clearly different from natural image captioning.
For instance, radiologists often describe abnormalities with precise clinical terms (\eg, “mild bibasal atelectatic opacities”) while using negation for normal findings (\eg, “no 
pneumothorax”). 
Many recent works~\cite{liu2024bootstrapping,che2025llm,li2024contrastive} exploit the strong language generation capabilities of LLMs to enhance model performance. However, without essential report knowledge, they treat all phrases/expressions in the report equal during learning. As clinically critical contents (\eg, 
abnormal findings) are rather sparsely distributed, the model learning would then be dominated by more frequent but less informative expressions (\eg, connective or comparative phrases like "compared with"). As a result, they may achieve strong performance on traditional image captioning metrics, while their improvement on clinical efficacy (CE) metrics remains limited.

To effectively leverage clinical knowledge in RRG task,
we are inspired by the recent development of multimodal large language models (MLLMs)~\cite{achiam2023gpt, touvron2023llama}: first, their visual grounding capability motivates us to extract discriminative visual anatomical features in a text-promptable way by leveraging the rich \textbf{anatomical knowledge} in MLLM; 
second, their text generation capability enables us to decode the extracted visual features auto-regressively into fluent and meaningful radiology reports; 
third, their multimodal aligning ability inspires us to perform disease-oriented clinical alignment across cases to integrate \textbf{disease knowledge} into the MLLM;
finally, their associated reinforcement learning from human feedback (RLHF) scheme~\cite{ouyang2022training} enlightens us to leverage \textbf{report knowledge} by further reinforcing the clinical quality of the generated report towards that of radiologists. 
Overall, we propose a novel \textbf{MLLM}-driven Chest X-ray \textbf{R}adiology \textbf{R}eport \textbf{G}eneration method with clinical knowledge enhancement, namely \textbf{MLLM-RRG}. As shown in Figure~\ref{fig:openfig} (b), it comprises four parts:

The first part is the \textit{referring anatomical feature extractor}. 
Instead of using a pretrained segmentation model or detector,  
we generate text descriptions of different anatomical regions (\eg left lung, spine) through an LLM (\eg, GPT-4 \cite{achiam2023gpt}) and directly interact them with the chest X-ray image to extract region features in the embedding space. This way guarantees a comprehensive coverage of the medical image. 

The second part is the \textit{multimodal report generator}, designed to generate accurate and comprehensive report from extracted visual features.
Benefiting from the inherent strong text generation capabilities of MLLMs~\cite{zhaommicl}, we design multimodal prompts that include dedicated visual features and textual instructions to guide the decoder of a MLLM (\eg, Multimodal Flan-T5~\cite{zhaommicl}) for report generation.

The third part is the \textit{clinical classification and alignment}. 
Within individual chest X-ray cases, we perform disease classification on chest X-ray images to strengthen disease-related information in the model. 
Across chest X-ray cases, we introduce a disease-oriented clinical alignment scheme: we first build a disease correlation matrix to model inter-disease relationship and incorporate it into the computation of disease similarity between cases. Based on this similarity measurement, each chest X-ray image is aligned to both its corresponding report and reports from other cases with relevant diseases, enabling cross-case disease knowledge transfer to produce more discriminative visual features.


Once the model is trained, in the fourth part, to leverage report knowledge, as per radiologists do, we include a \textit{clinical quality reinforcement learning} (CQRL) scheme to improve the clinical quality of the generated report. 
Since real radiologist’s feedback is difficult to achieve, we develop a surrogate feedback function from a metric, radiology reports clinical quality (RadCliQ)~\cite{yu2023evaluating},  to fit in the RLHF, which specifically encourages the model to focus on clinically critical content, eventually producing fluent, professional and clinically accurate radiology reports.

Overall, we summarize the contributions of this work: 
\begin{compactitem}
    \item From the technical side, we develop MLLM-RRG with dedicated multimodal instructions to guide MLLM to auto-regressively generate radiology reports from chest X-ray images. We offer several new components in this process, such as referring anatomical feature extractor, disease-oriented clinical alignment, clinical quality reinforcement learning, \etc, to inject three types of clinical knowledge, anatomical, disease and report knowledge into the MLLM to overall enhance the clinical significance of the generated reports.
    \item From the clinical side, we perform clinical test by inviting clinicians to re-annotate some disease labels for the main benchmark datasets and carefully evaluate the generated reports. Extensive experiments on MIMIC-CXR~\cite{johnson2019mimic} and IU X-Ray~\cite{demner2016preparing} datasets demonstrate that MLLM-RRG not only surpasses the state of the art on CE metrics but also achieves an average clinical score of 8.615 out of 10, indicating the generated reports well match the quality of those created by clinicians.
\end{compactitem}

\begin{figure*}[t]
\includegraphics[width=\textwidth]{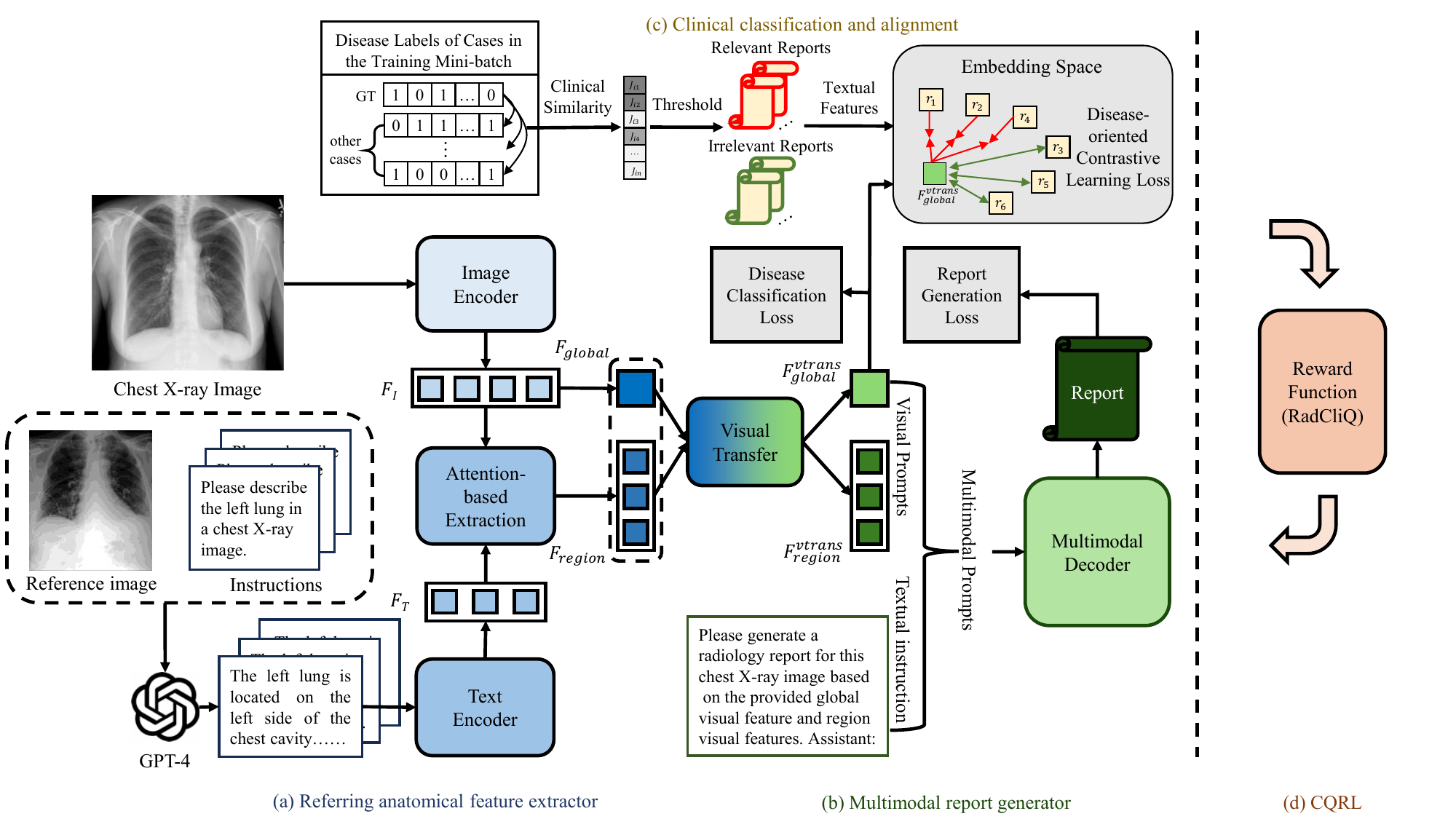}
\caption{The overall framework of our MLLM-RRG, which comprises four parts. a) Referring anatomical feature extractor: we leverage anatomical knowledge to extract referring visual anatomical features from different regions of the chest X-ray image.
b) Multimodal report generator: we design multimodal prompts that include visual features and textual instruction to guide a multimodal decoder to produce accurate and comprehensive report. 
c) Clinical classification and alignment: we first perform disease classification based on the extracted visual features; secondly,  introduce a disease-oriented clinical alignment scheme 
to integrate disease knowledge into the model by aligning visual features not only with their corresponding report, but also with reports from other patients exhibiting similar diseases. 
d) Clinical quality reinforcement learning (CQRL): we develop the RadCliQ reward function to leverage report knowledge for improving the clinical relevance between the generated and written reports.}
\label{fig:overview}
\end{figure*}

\section{Related Works}
\subsection{Radiology report generation}
Radiology report generation (RRG) aims to automatically generate descriptive diagnostic reports for given images. Early approaches adopt the CNN-RNN architectures, where CNNs are used to extract visual features, followed by RNNs for sequential text generation \cite{zhang2020radiology}. With the advent of Transformers, more recent methods exploit them to enhance cross-modal interactions and long-range dependency modelling \cite{chen2020generating,li2024contrastive}. For example, R2Gen \cite{chen2020generating}  utilizes a Transformer-based architecture to capture similar patterns across different X-ray images, enabling more coherent and structured report generation. Li \etal \cite{li2024contrastive} employ the LLM (\ie, GPT-2) as the decoder to generate the report. Beyond pure encoder-decoder designs, graph-based modelling has also been explored to incorporate medical knowledge or auxiliary modalities, such as encoding clinical history together with chest X-ray images and fusing them via a graph network~\cite{shang2022matnet}, or constructing a knowledge graph to inject diagnostic cues into visual representations~\cite{xiang2024gmod,daydar2026unsupervised}. In addition, hierarchical alignment \cite{li2023unify} and memory mechanism \cite{tao2024memory} have been explored to facilitate the generation of long, consistent, and clinically informative reports.

Recently, retrieval-based methods have gained increasing attention in RRG. These methods retrieve similar cases from the dataset and incorporate the retrieved information into either the decoding stage or the representation learning stage to enhance the report quality \cite{tao2024memory,liu2024bootstrapping,park2025dart,lin2025dc}. 
For instance, Tao \etal \cite{tao2024memory} retrieve images/reports from a memory bank with the query image or report; then align the corresponding retrieved visual or textural features in a shared embedding space via contrastive learning. 
Similarly, Liu \mbox{\etal} \cite{liu2024bootstrapping} adopt a contrastive objective that encourages higher feature similarity between the query image and its paired report while pushing it away from other reports. 
Lin \mbox{\etal} \mbox{\cite{lin2025dc}} build a multimodal retriever to fetch similar reports with the query image and feed the retrieved reports as auxiliary inputs into the decoder. 
However, these methods rely on image-to-image or image-to-text similarities in the retrieval process, making the retrieved reports may not truly be disease-relevant.
More recently, Park \mbox{\etal \cite{park2025dart}} introduce a disease-matching constraint to retrieve reports. Nevertheless, they strictly require that retrieved cases must exhibit exact diseases as the query.  
Consequently, clinically correlated cases may be mistakenly treated as irrelevant. 

In contrast, we propose a disease-oriented clinical alignment mechanism that incorporates inter-disease co-occurrence correlations into the computation of case similarity. By this means, 
cases sharing partially overlapping diseases can be effectively retrieved as long as they are clinically relevant
, allowing the model to exploit more diverse disease evidence.

\subsection{Multimodal large language model}
Inspired by the powerful text generation capabilities of LLMs, recent research has extended LLMs to the visual domain, leading to the emergence of multimodal large language models (MLLMs) \cite{li2023blip,liu2023visual,wang2024qwen2}. These models typically transform the visual representation from a vision encoder into the language space via a projection/alignment module, and subsequently leverage the LLM for semantic understanding and text generation. For instance, BLIP-2 \cite{li2023blip} employs a lightweight query-based transformer to bridge vision encoders and LLM, enabling efficient adaptation to many downstream multimodal tasks. LLaVA \cite{liu2023visual} integrates CLIP-based visual features into Vicuna \cite{chiang2023vicuna}, an LLM derived from LLaMA \cite{touvron2023llama}, via trainable projection layers. Qwen-VL \cite{wang2024qwen2} connects a Vision Transformer with the Qwen language model \cite{bai2023qwen} through cross-modal alignment modules, which are further optimized via instruction tuning for downstream tasks. 

Recently, several studies have applied MLLMs to the RRG task. For instance, Liu \etal \cite{liu2024bootstrapping} leverage the vision encoder and multimodal decoder of MiniGPT-4 \cite{zhu2023minigpt} for feature extraction and report generation. Hou \etal \cite{hou2025radar} adopt the pre-trained BLIP-3 \cite{xue2024xgen} as the backbone to improve the quality of the generated reports. 
The improvement in linguistic quality, however, does not necessarily lead to higher clinical accuracy in these works. We instead inject three types of clinical knowledge into the MLLM, enabling it to generate clinically accurate and professional reports.

\subsection{Reinforcement learning}
Reinforcement learning (RL) is a learning paradigm where an agent learns to make decisions by optimizing cumulative rewards through interaction with an environment \cite{sutton1998reinforcement}. RL has been increasingly utilized to help the generation models, particularly in scenarios where standard supervised learning objectives are hard to achieve \cite{kumar2022generative,nie2021triangle,lu2022good}. Tasks benefit from RL include machine translation \cite{kumar2022generative}, image captioning \cite{nie2021triangle}, visual question answering \cite{lu2022good}, \etc

Recently, RL has also been applied to the RRG task, providing additional supervision through carefully designed reward functions \cite{qin2022reinforced,jing2025reason}. 
For example, Qin \etal \cite{qin2022reinforced} leverage BLEU as a reward signal to guide the cross-modal alignment between visual and textual features.  
Jing \mbox{\etal} \mbox{\cite{jing2025reason}} adopt a text format-based reward and an IoU-based reward to encourage model reasoning and improve the semantic consistency between generated texts and anatomical regions respectively. 
However, these rewards are primarily designed to improve natural language generation (NLG) metrics and fail to enhance the generated reports from a clinical-grounded perspective, making the improvement on clinical efficacy metrics negligible. 
In this paper, we instead develop the radiology reports clinical quality (RadCliQ) reward function to help the model focus on improving the clinically critical parts of the generated reports.

\section{Method}

\begin{table*}[t]
    \centering
    \caption{The text descriptions of different anatomical regions generated by GPT-4. Due to page limitations, we only provide descriptions for 7 regions.}
    \label{tab:descriptions}

    \small
    \renewcommand{\arraystretch}{1.15}

    \begin{tabularx}{\textwidth}{
        >{\raggedright\arraybackslash}p{0.18\textwidth} |
        >{\raggedright\arraybackslash}X
    }
    \hline
    \textbf{Region} & \textbf{Description} \\
    \hline
    Left lung &
    The left lung presents as a transparent area with two lobes, bordered by the ribcage, extending from collarbone to diaphragm, adjacent to the heart’s left border, showing clear fields without opacities, masses, or fluid levels, indicating normal pulmonary status. \\
    \hline
    Right lung & The right lung, larger than the left, spans from collarbone to diaphragm, enclosed by the ribcage, divided into three lobes by distinct fissures, and displays clear, transparent lung fields without abnormalities, ensuring a healthy pulmonary condition adjacent to the heart’s edge. \\
    \hline
    Cardiac silhouette &
    The cardiac silhouette on a chest X-ray, appearing as a left located, denser area with smooth contours, occupies less than half of the thoracic width on the left side, providing insights into the heart’s condition through its shape and size, indicative of the individual’s normal cardiac health when aligned with expected standards for their body size and age. \\
    \hline
    Mediastinum & The mediastinum is visible as the central thoracic compartment with varied densities, housing critical structures between the lungs, marked by well-centered positioning, clear margins, and standard width; its normal appearance—free from widening, mass effects, or shifts—indicates the absence of underlying pathology and maintains thoracic health. \\
    \hline
    Right hilar structures & The right hilar structures, comprising bronchi, blood vessels, lymph nodes, and nerves, are centrally positioned on the right side, slightly lower than the left, displaying clear and defined edges without enlargement or abnormal shadows, indicating normal pulmonary and vascular health when devoid of disease indicators. \\
    \hline
    Left costophrenic angle & The left costophrenic angle, formed by the intersection of the diaphragm and left rib cage, appears as a sharp, clear space at the lung’s lower edge, crucial for evaluating pleural space and lung health; its well-defined visibility without blunting or fluid levels signals no pleural effusion, reflecting normal pulmonary status. \\
    \hline
    Spine & The spine appears as a straight, central column of increasing-sized rectangular vertebrae, properly aligned without misalignment or fractures, behind heart and mediastinal structures; uniform disc spaces without narrowing indicate the absence of degenerative changes, essential for assessing spinal integrity and overall thoracic health. \\
    \hline
    \end{tabularx}
\end{table*}

Given a chest X-ray image $I \in \mathbb{R}^{H \times W \times 3}$, where $H$ and $W$ is the height and width of the image, our method aims to generate the corresponding radiology report $R$. 
As shown in Figure~\ref{fig:overview}, our method includes four parts: referring anatomical feature extractor, multimodal report generator, clinical classification and alignment, and clinical quality reinforcement learning.
Below we detail them respectively.

\subsection{Referring anatomical feature extractor}
\label{sec:feature_extractor}
Radiologists leverage anatomical knowledge to systematically examine different anatomical regions when interpreting a chest X-ray image. Different from previous methods that rely on pre-trained segmentation models or region detectors~\cite{tanida2023interactive,gu2024complex}, we design a referring anatomical feature extractor to inject such anatomical knowledge into the model in a text-promptable way. The feature extractor embeds the image into visual features corresponding to 29 different anatomical regions (\eg, left lung, mediastinum, spine).
These regions are defined in~\cite{wu2021chest} by board-certified radiologists based on three principles: (1) clinical usage in radiology reports; (2) visual detectability on 2D chest X-ray images; and (3) richness in clinically meaningful attributes.

Specifically, we utilize text descriptions of these regions generated by a pretrained LLM (\eg GPT-4~\cite{achiam2023gpt}) to interact with the image (see Table~\ref{tab:descriptions} for generated descriptions). 
These text descriptions are generated by prompting the LLM with a reference image randomly sampled from the dataset and the instruction ``Please describe the [class] in a chest X-ray'', where the [class] token is replaced with the region name. 
These descriptions are denoted by $T \in \{t_{k} | k=1, 2, ..., K\}$ for $K$ (\ie 29) different regions.
Unlike region detector-based pipelines, the LLM is not part of the training or inference loop: it is used once to generate a fixed set of region-wise anatomical descriptions, which are then stored and treated as textual priors.
Then we employ the encoders from a medical CLIP model pre-trained on 2D medical image-text datasets~\cite{wang2022medclip}: its text encoder is used to extract the textual feature $F_{T}^{k} \in \mathbb{R}^{1 \times D}$ for each region description $t_k$, where $D$ is the dimension of the feature;
its image encoder is used to extract the visual features $F_{I} \in \mathbb{R}^{N \times D}$ from the chest X-ray image $I$, where $N$ represents the number of visual tokens.
Below, we specify the extraction of the referring visual anatomical features given $F_{I}$ and $F_{T}^{k}$.

\myparagraph{Referring visual anatomical features}
We propose to inject anatomical knowledge (text descriptions corresponding to 29 anatomical regions) through a referring way, avoiding explicit anatomical region detection.
An attention mechanism is employed to extract the visual anatomical features $F_{region}^{k}$ from the visual features $F_I$ through the guidance of textual features $F_T^k$.
Initially, we use \emph{cross-attention}, treating $F_I$ as the query and $F_{T}^{k}$ as both key and value, to facilitate their interaction. This process let $F_I$ perceive visual information pertinent to the textual information of the corresponding anatomical region.
It is expressed as:
$F_{CA}^k = \text{CA}(\text{LN}(F_I), F_{T}^{k}) + F_I$,
where $\text{CA}(\cdot)$ represents the cross-attention operation and $LN(\cdot)$ the layer normalization. 
Subsequently, we apply a self-attention to $F_{CA}^k$ for enhancing its region-specific information.
This step is described as:
$F_{SA}^k = \text{SA}(\text{LN}(F_{CA}^k)) + F_{CA}^k$, 
where $\text{SA}(\cdot)$ represents self-attention operation.
Finally, we introduce a feed-forward network to encode and further refine $F_{SA}^k$, denoted as: 
$F_{FFN}^k = \text{FFN}(\text{LN}(F_{SA}^k)) + F_{SA}^k$,
where $\text{FFN}(\cdot)$ denotes feed-forward network.
Note, $\text{CA}(\cdot)$, $\text{SA}(\cdot)$, and $\text{FFN}(\cdot)$ are components within a block. By stacking multiple such blocks (\ie 3 blocks) and applying mean pooling along the token dimension, we ultimately obtain a compact anatomical region representation $F_{region}^k \in \mathbb{R}^{1 \times D}$.
By processing all $K$ anatomical regions in parallel and concatenating their outputs, we obtain the final visual anatomical features $F_{region} \in \mathbb{R}^{K \times D}$. 

Considering that the image's global information can also benefit the model for report generation, we further extract the visual global feature.
To achieve this, we simply take the average of the $N$ extracted patch features from the original visual features $F_{I} \in \mathbb{R}^{N \times D}$ along the token dimension and obtain the visual global feature, denoted by $F_{global} \in \mathbb{R}^{1 \times D}$.

\subsection{Multimodal report generator}
\label{sec:report_generator}

After extracting visual features $F_{global}$ and $F_{region}$, we aim to produce accurate and comprehensive radiology report $R$.
To achieve it, we first construct multimodal prompts, which comprise  $F_{global}$ and $F_{region}$, as well as the textual instruction.
These multimodal prompts are then fed into a multimodal decoder to auto-regressively generate the radiology report.

\myparagraph{Multimodal prompts construction}
The multimodal prompts consist of two components: the visual prompts (global and local features) and the textual prompts (instruction).
We need to process them to ensure they can be compatible with the multimodal decoder's input format.

\noindent \emph{Visual prompts.} $F_{global}$ and $F_{region}$ cannot be directly fed into the multimodal decoder. Motivated by \cite{li2023blip}, we construct a visual transfer module $\text{VTrans}(\cdot)$ to transform these features into ones that can be fed into the decoder.
Unlike ~\cite{li2023blip}, which compresses large number of visual features into a small number through learnable queries, our visual features are not many, hence do not need to be compressed.
Therefore, we employ a standard 3-layer transformer that each layer includes only a self-attention and a feed-forward network as the visual transfer module $\text{VTrans}(\cdot)$.
This module inputs $F_{global}$ and $F_{region}$ and outputs their transferred versions, formulated as
\begin{equation}
    F_{global}^{vtrans}, F_{region}^{vtrans} = \text{VTrans}(F_{global}, F_{region}).
\end{equation}


\noindent \emph{Textual prompts.} For textual prompts, the aim is to ensure that the multimodal decoder understands the meaning of given features and what actions are required based on these features.
Following this principle, we construct instruction $T_{inst}$, \ie ``Please generate a radiology report for this chest X-ray image based on the provided visual global feature and visual anatomical features. Assistant:~''.
We then use the same text encoder in Sec.~\ref{sec:feature_extractor} to transform this instruction into features $F_{inst} \in \mathbb{R}^{L \times D}$, where $L$ represents the length of the instruction and $D$ denotes the feature dimension.

We concatenate the features corresponding to the visual and textual prompts along the token dimension to obtain the multimodal prompts, $F_{multimodal} \in \mathbb{R}^{(1+K+L) \times D}$, which are then input into the multimodal decoder.

\myparagraph{Multimodal decoder}
After the construction of the multimodal prompts, we input these prompts into the multimodal decoder of a MLLM, \ie Multimodal Flan-T5~\cite{zhaommicl}, and adopt an auto-regressive way for the generation of the radiology report $R$.
Specifically, for the $i$-th token $t_i$ to be generated in the report, prediction is based on the multimodal prompts and all previously generated tokens, expressed as:
\begin{equation}
    t_i = \phi(F_{multimodal}, t_0, \cdots, t_{i-1}), i \in \{0, 1, \cdots, M\},
\end{equation}
where $M$ is the maximum length of the generated report and $\phi(\cdot)$ is the multimodal decoder.
The auto-regressive prediction proceeds until the token indicates the end of report (\ie [EOS]), ultimately yielding the radiology report $R$. 

\subsection{Clinical classification and alignment}
\label{sec:cla}
Radiologists utilize 
prior diagnostic experience to support the interpretation of new cases. To integrate such disease knowledge into our model, we design a clinical classification and alignment scheme.
Given the visual global feature $F_{global}^{vtrans} \in \mathbb{R}^{1 \times D}$, we follow \cite{park2025dart} to apply a linear layer for clinical classification.
Moreover, we introduce a disease-oriented clinical alignment scheme by aligning $F_{global}^{vtrans}$ not only with the textual features of the corresponding ground-truth report but also with those of clinically relevant reports from other patients.


\begin{figure}[t]
    \centering
    \includegraphics[width=\columnwidth]{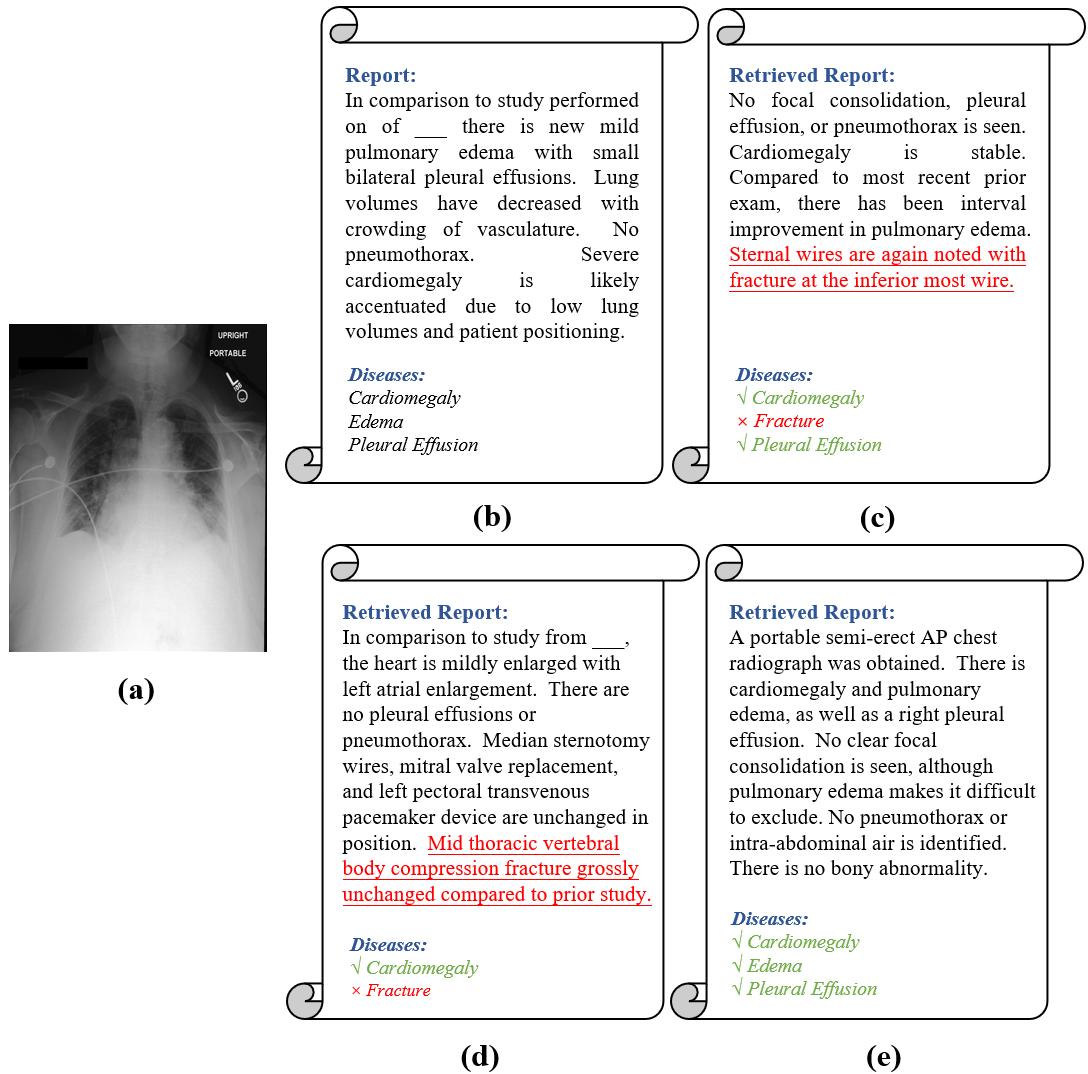}
    \caption{(a) Chest X-ray image. (b) Ground-truth report. (c), (d) and (e) show retrieved reports based on image-to-image similarity, image-to-text similarity and our proposed disease-oriented clinical alignment scheme, respectively. Inconsistent sentences/disease are highlighted in red. Compared with other methods, our scheme retrieves reports that exhibit higher clinical relevance with the ground truth.}
    \label{fig:disease_oriented_clinical_alignment}
\end{figure}

\myparagraph{Disease-oriented clinical alignment scheme}
We use the same text encoder described in Sec.~\ref{sec:feature_extractor} to extract textual features $C=\left \{c_1,...,c_m,...,c_{B}  \right \}$ from reports of cases 
within the training mini-batch, where $c_1\in \mathbb{R}^{1\times D}$ and $B$ denotes the batch size. Assuming $c_m$ corresponds to the textual feature of ground truth report in 
the $m$-th case, we first follow the common practice in contrastive learning by aligning it with the visual global feature of the $m$-th case, \ie maximizing the cosine similarity between $c_m$ and $F_{global}^{vtrans}$. Next, based on the case disease labels, we retrieve cases that are clinically relevant to the $m$-th case. 

We first incorporate the correlations among diseases into the retrieval process 
to allow cases with clinically relevant diseases 
can be effectively retrieved. 
For example, a case presenting consolidation can be matched with one presenting both consolidation and pneumonia, as consolidation is a common manifestation of pneumonia \cite{kayser2022explaining}.
Specifically, we compute a disease correlation matrix $C \in \mathbb{R}^{D\times D}$ from the dataset: 
\begin{equation}
    C_{ij} = \frac{k_{ij}}{\min(k_i, k_j)}
    \label{eqn:3}
\end{equation}
where $k_{ij}$ denotes the number of cases in which diseases $i$ and $j$ co-occur, and $k_i$, $k_j$ are the numbers of cases in which disease $i$ and disease $j$ appear, respectively. 
Then, inspired by the standard Tanimoto coefficient~\cite{rogers1960computer}, we define a disease-aware Tanimoto similarity between two cases $m,n$ with disease labels $v,t \in \{0,1\}^D$ as
\begin{equation}
    T(m,n)=\frac{v^TCt}{v^TCv+t^TCt-v^TCt+\epsilon}
\end{equation}
where $\epsilon$ is a small constant for numerical stability; $D$ is the length of a disease label. In this way, the correlation between diseases is incorporated into the similarity computation.
We then select the cases whose clinical relevance 
with the $m$-th case exceed a predefined threshold $\tau$. We denote the indices of these cases in the mini-batch as 
$\mathcal{P}_m = \left\{ n \mid T(m,n) > \tau, \ n\in \left \{ 1,2,...,B \right \} \right\}$
The textual features of these cases are then jointly aligned with the visual global feature $F_{global}^{vtrans}$ of the $m$-th case, while the textual features of the remaining cases in the mini-batch are treated as negatives in disease-oriented contrastive learning loss $\mathcal{L}_{con}$:
\begin{equation}
\mathcal{L}_{\text{con}} = - \frac{1}{|\mathcal{P}_i|} \sum_{j \in \mathcal{P}_i} \log \frac{\exp\left( \text{sim}(F_{global}^{vtrans}, c_j)/\mu   \right)}{\sum_{k=1}^{n} \exp\left( \text{sim}(F_{global}^{vtrans}, c_k)/\mu   \right)}
\label{eq:5}
\end{equation}
where $sim(.,.)$ denotes the cosine similarity, and $\mu$ is the temperature. If no relevant cases are found (\ie $\mathcal{P}_i = \emptyset$), we only align $F_{global}^{vtrans}$ of the $m$-th case with its ground-truth report's textual feature $c_m$.

As shown in Figure~\ref{fig:disease_oriented_clinical_alignment}, compared with previous methods that retrieve cases based on image-to-image or image-to-text similarities, our disease-oriented clinical alignment scheme retrieves more clinically relevant cases, thereby helping the model extract more discriminative visual features and achieve better alignment with disease semantics.

\subsection{Clinical quality reinforcement learning}
\label{sec:RL}
Radiologists use report knowledge to write reports that are both clinically accurate and professionally standardized. 
Similarly,  AI model also needs the report knowledge to transform extracted visual features into high-quality radiology reports.
Therefore, 
we employ a clinical quality reinforcement learning (CQRL) scheme to further refine the report generator $G(\cdot)$ (\ie the visual transfer module and the multimodal decoder in Sec.~\ref{sec:report_generator}) after completing the supervised learning as above (see also Sec.~\ref{sec:model_training}). During the RL process, the parameters of other components in our MLLM-RRG are frozen.

Conventional evaluation metrics (\eg BLEU~\cite{papineni2002bleu}) fail to adequately
reflect the clinically significant errors (\ie, errors that might change diagnosis, assessment, or patient management~\cite{yu2023evaluating}) in the generated reports.
To address it, we employ the RadCliQ~\cite{yu2023evaluating}, which serves as a surrogate feedback for the radiologists, as the reward function $Reward(\cdot)$ in the RL process.
RadCliQ is composed of four metrics, including BLEU~\cite{papineni2002bleu}, BertScore~\cite{zhang2019bertscore}, CheXbert vector similarity~\cite{smit2020chexbert}, and RadGraph F1~\cite{jain2021radgraph}, and employs a fitting network over them to achieve the optimal combination. 
In this way, the model is explicitly guided to reduce clinically significant errors during RL, thereby improving the clinical quality of the generated reports.


Similar to the proximal policy optimization (PPO)~\cite{schulman2017proximal} used in RLHF~\cite{ouyang2022training}, we also use PPO to update model parameters.
First, we need to duplicate the report generator $G(\cdot)$ by creating two versions, one is the reference report generator $G_{ref}(\cdot)$ whose parameters are frozen, and the other is the model updated through PPO, defined as $G_{PPO}(\cdot)$.
We employ two losses, the reward loss to improve the model and the KL divergence loss to ensure that the outputs of two models do not diverge significantly.
\begin{gather}
    \mathcal{L}_{reward} = Reward(R, \hat{R}) \\
    \mathcal{L}_{KL} = Dist_{KL}(G_{ref}(E(I, T)), G_{PPO}(E(I, T)))
\end{gather}
where $Dist_{KL}$ is the operator to calculate the KL distance.
Note that lower RadCliQ indicates better results. Therefore, minimizing $\mathcal{L}_{reward}$ drives the generated report toward lower RadCliQ values and better clinical quality.
The total loss for the CQRL is
\begin{equation}\label{equ8}
    \mathcal{L}_{RL} = \mathcal{L}_{reward} + \lambda_{KL} \mathcal{L}_{KL}
\end{equation}
where $\lambda_{KL}$ is an adaptive weighting coefficient, and is adjusted according to the following formula:
\begin{equation}\label{eq:9}
\lambda_{KL} =
\begin{cases}
1, &Reward(R,\hat{R}) \le \theta,\\
Reward(R,\hat{R}) \times \theta, & others.
\end{cases}
\end{equation}
where $\theta$ acts as a controlling threshold.
When the model achieves a reward smaller than or equal to $\theta$, it indicates that the current generated report is close to the ground truth report.
In this case, by setting $\lambda_{KL} = 1$, we ensure that the model's output should be as close as possible to that of the reference model.
Conversely, if the gap between the model's generated report and the ground truth is significant, the loss should be turned to the reward loss.

\subsection{Model training}
\label{sec:model_training}
There are two learning stages for our method, the first is supervised learning and the second is the reinforcement learning.
For the first stage, all modules of our model are trained under a multitask learning framework, which includes three losses:
the language modelling loss $\mathcal{L}_{report}$~\cite{achiam2023gpt} to optimize the generated report, the disease classification loss $\mathcal{L}_{disease}$~\cite{park2025dart} to strengthen the visual features with more disease-related information, and the disease-oriented contrastive learning loss $\mathcal{L}_{con}$~(Eqn.~\ref{eq:5}) to align the visual features with the textual features of both its own report and clinically relevant reports from other cases. Specifically, $\mathcal{L}_{report}$ is computed upon the output report, and $\mathcal{L}_{disease}$ is applied to the visual global features for multi-class prediction (see Figure~\ref{fig:overview}). Both of $\mathcal{L}_{report}$ and $\mathcal{L}_{disease}$ use the cross-entropy form. The total loss $\mathcal{L} = \mathcal{L}_{report} + \lambda_1 \mathcal{L}_{disease} + \lambda_2 \mathcal{L}_{con}$, where $\lambda_1$ and $\lambda_2$ are loss weights. For the second stage, please refer to Sec.~\ref{sec:RL}.
\section{Experiments}

\begin{table*}[htbp]\small
\centering
\caption{Comparison of our method with state of the art on the MIMIC-CXR.}
\begin{tabular}{l|c|ccc|ccc|c}
\hline
\multicolumn{1}{c|}{\multirow{2}{*}{Method}} & \multicolumn{1}{c|}{\multirow{2}{*}{Venue}}  & \multicolumn{3}{c|}{NLG} & \multicolumn{3}{c|}{CE} & \multicolumn{1}{c}{\multirow{2}{*}{RadCliQ$\downarrow$}} \\
\cline{3-8} 
& & BLEU4$\uparrow$ & METEOR$\uparrow$ & ROUGE$\uparrow$ & Precision$\uparrow$ & Recall$\uparrow$ & F1$\uparrow$ & ~ \\ \hline
DCL \cite{li2023dynamic} & CVPR'23 & 0.109 & 0.150 & 0.284 & 0.471 & 0.352 & 0.373 & 1.449 \\
TSGET \cite{yi2024tsget} & JBHI'24 & 0.121 & 0.149 & 0.281 & 0.319 & 0.509 & 0.393 & -- \\
PromptMRG \cite{jin2024promptmrg} & AAAI'24 & 0.112 & 0.157 & 0.268 & 0.501 & 0.509 & 0.476 & 1.229 \\
COMG \cite{gu2024complex} & WACV'24 & 0.124 & 0.128 & 0.290 & 0.424 & 0.291 & 0.345 & 1.323 \\
GMoD \cite{xiang2024gmod} & MICCAI'24 & 0.124 & 0.166 & 0.286 & 0.457 & 0.353 & 0.399 & 1.169 \\
ECRG \cite{hou2024energy} & MICCAI'24 & 0.128 & 0.161 & 0.269 & 0.456 & 0.513 & 0.482 & -- \\
ORID \cite{gu2025orid} & WACV'25 & 0.117 & 0.150 & 0.284 & 0.435 & 0.295 & 0.352 & -- \\
LHR-RLF \cite{yi2024lhr} & TMI'25 & 0.120 & 0.154 & 0.296 & 0.392 & 0.335 & 0.342 & -- \\
STREAM \cite{yang2025spatio} & TMI'25 & \textbf{0.139} & 0.172 & 0.297 & 0.515 & 0.447 & 0.478 & -- \\
Sha \etal \cite{sha2025contrastive} & MICCAI'25 & 0.136 & 0.169 & \textbf{0.303} & 0.503 & 0.442 & 0.469 & -- \\
Diff-RRG \cite{yun2025diff} & MICCAI'25 & 0.120 & 0.164 & 0.276 & 0.528 & 0.430 & 0.474 & 1.140 \\
RIHA \cite{chen2026riha} & JBHI'26 & 0.109 & 0.159 & 0.293 & 0.486 & 0.298 & 0.375 & -- \\
MARE \cite{gao2026mare} & AAAI'26 & 0.133 & 0.161 & 0.291 & 0.433 & 0.378 & 0.375 & -- \\
Wu \etal \cite{wu2026disease} & AAAI'26 & 0.131 & -- & 0.291 & -- & -- & -- & -- \\
FedMRG \cite{che2025llm} & TMI'26 & 0.127 & -- & 0.276 & 0.454 & 0.302 & 0.339 & -- \\
\hline
\textbf{MLLM-RRG(Ours)} & -- & 0.131 & \textbf{0.174} & 0.299 & \textbf{0.529} & \textbf{0.550} & \textbf{0.539} & \textbf{0.825} \\
\hline
\end{tabular}
\label{tab:mimic-cxr}
\end{table*}

\begin{table*}[htbp]\small
\centering
\caption{Comparison of our method with state of the art on the IU X-Ray.}
\begin{tabular}{l|c|ccc|ccc|c}
\hline
\multicolumn{1}{c|}{\multirow{2}{*}{Method}} & \multicolumn{1}{c|}{\multirow{2}{*}{Venue}}  & \multicolumn{3}{c|}{NLG} & \multicolumn{3}{c|}{CE} & \multicolumn{1}{c}{\multirow{2}{*}{RadCliQ$\downarrow$}} \\
\cline{3-8} 
& & BLEU4$\uparrow$ & METEOR$\uparrow$ & ROUGE$\uparrow$ & Precision$\uparrow$ & Recall$\uparrow$ & F1$\uparrow$ & ~ \\ \hline
DCL \cite{li2023dynamic} & CVPR'23 & 0.163 & 0.193 & 0.383 & 0.721 & 0.650 & 0.684 & 0.719 \\
TSGET \cite{yi2024tsget} & JBHI'24 & 0.194 & 0.218 & 0.402 & -- & -- & -- & -- \\
PromptMRG \cite{jin2024promptmrg} & AAAI'24 & 0.098 & 0.160 & 0.281 & 0.490 & 0.389 & 0.434 & 0.979 \\
COMG \cite{gu2024complex} & WACV'24 & 0.206 & 0.218 & 0.383 & -- & -- & -- & -- \\
GMoD \cite{xiang2024gmod} & MICCAI'24 & 0.203 & 0.217 & \textbf{0.418} & -- & -- & -- & -- \\
ORID \cite{gu2025orid} & WACV'25 & 0.198 & 0.211 & 0.400 & -- & -- & -- & -- \\
LHR-RLF \cite{yi2024lhr} & TMI'25 & 0.210 & 0.223 & 0.416 & 0.722 & 0.677 & 0.699 & 0.819 \\
Sha \etal \cite{sha2025contrastive} & MICCAI'25 & 0.184 & 0.211 & 0.385 & -- & -- & -- & -- \\
STREAM \cite{yang2025spatio} & TMI'25 & 0.188 & 0.215 & 0.387 & -- & -- & -- & -- \\
RIHA \cite{chen2026riha} & JBHI'26 & 0.218 & 0.204 & 0.376 & -- & -- & -- & -- \\
Wu \etal \cite{wu2026disease} & AAAI'26 & 0.187 & -- & 0.384 & -- & -- & -- & -- \\
FedMRG \cite{che2025llm} & TMI'26 & 0.149 & -- & 0.336 & 0.162 & 0.154 & 0.154 & -- \\

\hline
\textbf{MLLM-RRG(Ours)} & -- & \textbf{0.235} & \textbf{0.230} & {0.417} & \textbf{0.752} & \textbf{0.687} & \textbf{0.718} & \textbf{0.501} \\
\hline
\end{tabular}
\label{tab:iu-xray}
\end{table*}

\subsection{Datasets and metrics}

\myparagraph{Datasets}
Following previous works \cite{chen2022cross,chen2020generating,qin2022reinforced}, we conduct our experiments on two widely used datasets: MIMIC-CXR \cite{johnson2019mimic} and IU X-Ray \cite{demner2016preparing}. MIMIC-CXR is the largest public dataset for radiology report generation, including 473,057 chest X-ray images and 206,563 corresponding reports from 63,478 patients at the Beth Israel Deaconess Medical Center. To ensure a fair comparison, we adopt the official data split, which includes 276,778 cases divided into 270,790 for training, 2,130 for validation, and 3,858 for testing. For IU X-Ray, we follow the data split protocol proposed by Chen \etal \cite{chen2020generating}. The 2,955 cases in the IU X-Ray dataset are split into 2,069, 296, and 590 for training, validation, and testing respectively.

Both datasets contain radiology images from multiple views (\ie frontal and lateral) along with the corresponding radiology reports. For MIMIC-CXR, we maintain a one-to-one correspondence between each image and its associated report. For IU X-Ray, we follow the common practice \cite{chen2020generating,chen2022cross,qin2022reinforced} to combine the frontal and lateral images from the same patient into a dual-channel input.



\myparagraph{Metrics}
We utilize metrics such as the Natural Language Generation (NLG), encompassing BLEU~\cite{papineni2002bleu}, METEOR~\cite{banerjee2005meteor}, and ROUGE~\cite{lin2004rouge}; Clinical Efficacy (CE), including precision, recall, and F1 scores; and Radiology Report Clinical Quality (RadCliQ)~\cite{yu2023evaluating}, which is a composite metric that includes BLEU~\cite{papineni2002bleu}, BERTScore~\cite{zhang2019bertscore}, CheXbert vector similarity~\cite{smit2020chexbert}, and RadGraph F1~\cite{jain2021radgraph}.
Note that for the NLG and CE metrics, higher values indicate better performance, whereas for RadCliQ, lower values signify better results.



\subsection{Implementation details}
Following previous works \cite{chen2022cross, chen2020generating, qin2022reinforced}, the original chest X-ray images from the MIMIC-CXR \cite{johnson2019mimic} and IU X-Ray \cite{demner2016preparing} datasets are resized to 256 $\times$ 256, randomly cropped to 224 $\times$ 224, and normalized to the range $[0,1]$. The reports are truncated to 100 and 60 words for MIMIC-CXR and IU X-Ray, respectively, with words occurring less than three times and irrelevant symbols removed~\cite{chen2020generating}. For data augmentation, we follow \cite{chen2022cross, chen2020generating, qin2022reinforced} to apply horizontal flip to each CXR image. For hyperparameters, we set the similarity threshold $\tau$ in Sec. \ref{sec:cla} to 0.60. The temperature $\mu$ in Eqn. \ref{eq:5} is set to 0.07. The controlling threshold $\theta$ in Eqn. \ref{eq:9} is set to 0.4. The loss weight $\lambda_1$ in Sec. \ref{sec:model_training} is set to 1.0 while $\lambda_2$ is set 
to 0.5. Above hyperparameters are determined based on the validation sets of datasets.

We utilize the encoders from MedCLIP \cite{wang2022medclip} as our image and text encoders. For the multimodal decoder, we utilize the decoder of Multimodal Flan-T5~\cite{zhaommicl}. The sizes of our encoders and decoder are identical (\mbox{\ie} 768). The training process is conducted on a Nvidia A40 GPU with a batch size of 32. The model is trained for 20 epochs in the first stage and 10 epochs in the second stage. For optimization, we use AdamW optimizer with a learning rate of $1\times10^{-4}$ and a weight decay of $5\times10^{-2}$. A combination of linear warm-up and cosine annealing scheduling is applied throughout the training.

\subsection{Comparison to state of the art} 
\label{sec::model_comparison}
\myparagraph{MIMIC-CXR}
Table~\ref{tab:mimic-cxr} shows the results of our MLLM-RRG and previous methods on the MIMIC-CXR dataset.
Compared with the previous best-performing method on the NLG metrics, \ie STREAM~\cite{yang2025spatio} and Sha \etal \cite{sha2025contrastive}, our method significantly outperforms the aforementioned two methods on the CE metrics, \eg +0.061 and +0.070 in F1, respectively.
On the other hand, compared to the previous best-performing method on the CE metrics, \ie ECRG \cite{hou2024energy}, our method significantly surpasses it by 0.057 in F1 and 0.030 in ROUGE. 
In terms of the RadCliQ metric, our MLLM-RRG, benefiting from using RadCliQ as the feedback function, shows a substantial improvement over previous methods.
These results demonstrate the effectiveness of our proposed MLLM-driven radiology report generation method. 



\myparagraph{IU X-Ray}
Table~\ref{tab:iu-xray} shows the performance comparison between our method and state of the art on the IU X-Ray dataset. 
The results of our method exceeds state of the art on both NLG and CE metrics 
(\eg it increases 0.025/0.019 in BLEU4/F1 from the very recent work LHR-RLF \cite{yi2024lhr}).

\myparagraph{Training cost and Model complexity}
Table~\ref{tab:training_efficiency} shows the comparison on training cost and model complexity between our method and state of the art. It’s worth noting that our model has the lowest parameters, FLOPs and memory consumption among all comparable LLM-based approaches, because we use a very light-weighted LLM, Flan-T5. Note that the parameters and FLOPs of 
RIHA~\cite{chen2026riha}, Sha \etal~\cite{sha2025contrastive}, MARE~\cite{gao2026mare}, Wu \etal \cite{wu2026disease} and FedMRG~\cite{che2025llm} are estimated by ourselves due to the absence of open-sourced codes.

\begin{table}[htbp]\small
\centering
\caption{Comparison of our method with state of the art on complexity and efficiency. }
\begin{tabular}{l|c|ccc}
\hline
Methods & LLM & Params$\downarrow$ & FLOPs$\downarrow$ & Mem.$\downarrow$ \\ \hline
DCL \cite{li2023dynamic} & \XSolidBrush & 644M & 523G & 1.2GB \\
PromptMRG \cite{jin2024promptmrg} & \XSolidBrush & 220M & 122G & 0.8GB \\
COMG \cite{gu2024complex} & \XSolidBrush & 83.2M & 59.9G & 0.7GB \\
GMoD \cite{xiang2024gmod} & \XSolidBrush & 26.1M & 9.60G & 0.2GB \\
LHR-RLF \cite{yi2024lhr} & \XSolidBrush & 82.1M & 1.47T & 2.0GB \\
RIHA \cite{chen2026riha} & \XSolidBrush & 190M & 180G & -- \\
Sha \etal \cite{sha2025contrastive} & \Checkmark & 8.4B & 10T & -- \\
Diff-RRG \cite{yun2025diff} & \Checkmark & 7.4B & 22.4T & 4.0GB \\
STREAM \cite{yang2025spatio} & \Checkmark & 1.21B & 2.09T & 2.5GB \\
MARE \cite{gao2026mare} & \Checkmark & 6.9B & 10T & -- \\
Wu \etal \cite{wu2026disease} & \checkmark & 14.2B & 25T & -- \\
FedMRG \cite{che2025llm} & \Checkmark & 6.8B & 10T & -- \\
\hline
\textbf{MLLM-RRG(Ours)} & \Checkmark & 542M & 276G & 1.3GB \\ \hline
\end{tabular}
\label{tab:training_efficiency}
\end{table}

\subsection{Ablation study}\label{sec:abalation}
We follow previous methods~\cite{chen2020generating,yang2025spatio} to conduct ablation study on MIMIC-CXR.


\begin{table*}[htbp]\small
    \centering
    \caption{Ablation study for referring anatomical feature extractor, multimodal report generator, clinical classification and alignment, and clinical quality reinforcement learning.}
    \begin{tabular}{l|ccc|ccc|c}
    \hline
    \multicolumn{1}{c|}{\multirow{2}{*}{Ablation}}&\multicolumn{3}{c|}{NLG} & \multicolumn{3}{c|}{CE} & \multicolumn{1}{c}{\multirow{2}{*}{RadCliQ$\downarrow$}} \\
    \cline{2-7}
    ~ & BLEU4$\uparrow$ & METEOR$\uparrow$ & ROUGE$\uparrow$ & Precision$\uparrow$ & Recall$\uparrow$ & F1$\uparrow$ & ~ \\ \hline
    MLLM-RRG & 0.131 & 0.174 & 0.299 & 0.529 & 0.550 & 0.539 & 0.825 \\ 
    \hdashline
    w/o $F_{global}$ & 0.124 & 0.168 & 0.294 & 0.504 & 0.542 & 0.522 & 0.855 \\        
    w/o $F_{region}$ & 0.118 & 0.159 & 0.291 & 0.476 & 0.509 & 0.492 & 1.051 \\        
    text-guided $\rightarrow$ explicit detection & 0.127 & 0.165 & 0.292 & 0.497 & 0.528 & 0.512 & 0.903 \\ \hdashline         
    descriptions: LLM $\rightarrow$ names & 0.120 & 0.158 & 0.294 & 0.475 & 0.488 & 0.481 & 1.120 \\
    description: LLM $\rightarrow$ random phrases & 0.122 & 0.160 & 0.288 & 0.490 & 0.494 & 0.492 & 1.104 \\
    description: LLM $\rightarrow$ one sentence & 0.125 & 0.172 & 0.295 & 0.511 & 0.540 & 0.525 & 0.854 \\
    description: with subtle bias & 0.123 & 0.168 & 0.295 & 0.495 & 0.502 & 0.498 & 1.079 \\
    description: LLM $\rightarrow$ clinician & 0.130 & 0.173 & 0.298 & 0.527 & 0.549 & 0.538 & 0.825 \\ \hdashline 
    LLM: GPT $\rightarrow$ Claude & 0.128 & 0.173 & 0.298 & 0.524 & 0.549 & 0.536 & 0.827 \\
    LLM: GPT $\rightarrow$ Deepseek & 0.129 & 0.173 & 0.296 & 0.525 & 0.546 & 0.535 & 0.827 \\
    LLM: text only& 0.128 & 0.171 & 0.296 & 0.516 & 0.541 & 0.528 & 0.834 \\ \hdashline
    ref image: normal & 0.131 & 0.174 & 0.298 & 0.528 & 0.549 & 0.538 & 0.826 \\
    ref image: abnormal & 0.129 & 0.174 & 0.298 & 0.526 & 0.549 & 0.537 & 0.828 \\
    ref image: fixed $\rightarrow$ specific & 0.130 & 0.172 & 0.298 & 0.517 & 0.549 & 0.532 & 0.831 \\ \hdashline
    w/o $F_{inst}$ & 0.122 & 0.166 & 0.292 & 0.488 & 0.536 & 0.511 & 0.911 \\
    w/o VTrans & 0.122 & 0.168 & 0.296 & 0.511 & 0.537 & 0.524 & 0.853 \\ \hdashline
    w/o $\mathcal{L}_{disease}$ & 0.125 & 0.166 & 0.293 & 0.471 & 0.492 & 0.481 & 1.040 \\
    w/o DOCA & 0.126 & 0.167 & 0.294 & 0.482 & 0.522 & 0.501 & 0.897 \\     
    DOCA $\rightarrow$ IIS & 0.127 & 0.172 & 0.296 & 0.499 & 0.534 & 0.516 & 0.869 \\    
    DOCA $\rightarrow$ IRS & 0.126 & 0.168 & 0.295 & 0.492 & 0.531 & 0.511 & 0.875 \\    
    DOCA: w/o disease matrix & 0.128 & 0.172 & 0.300 & 0.512 & 0.549 & 0.530 & 0.830 \\
    DOCA: batch $\rightarrow$ dataset & 0.129 & 0.170 & 0.296 & 0.517 & 0.546 & 0.531 & 0.831 \\ \hdashline
    w/o CQRL & 0.125 & 0.170 & 0.296 & 0.518 & 0.540 & 0.529 & 1.018  \\  
    reward: RadCliQ $\rightarrow$ BLEU4 & 0.136 & 0.163 & 0.292 & 0.488 & 0.521 & 0.504 & 1.116  \\
    w/ fixed $\lambda_{KL}$ & 0.126 & 0.172 & 0.295 & 0.510 & 0.543 & 0.526 & 0.901  \\
    \hline
    \end{tabular}
    \label{tab:ablation}
\end{table*}

\noindent \textbf{Referring anatomical feature extractor.}
We study the impact of visual anatomical and global features, as well as the text descriptions used to inject anatomical knowledge. 

\noindent \emph{Visual global and anatomical features.} To validate the effect of using both visual global and anatomical features, we conduct experiments by respectively removing them from the feature extractor.
In Table~\ref{tab:ablation}, the results show that when global feature is removed (w/o $F_{global}$), there is a performance decline on all metrics, validating the contribution of the global information to report generation.
On the other hand, when we remove the visual anatomical features (w/o $F_{region}$), we observe a more significant decrease on all metrics (\eg -0.053 in precision and -0.047 in F1), indicating that visual anatomical features are crucial for accurate report generation.

If we replace this implicit referring strategy with explicit region detection using Faster R-CNN detector \cite{ren2015faster} (\ie text-guided $\rightarrow$ explicit detection in Table~\ref{tab:ablation}), the F1 decreases from 0.539 to 0.512, indicating the effectiveness of our strategy.

\noindent \emph{Text prompts for visual anatomical features.}
To validate the effect of obtaining text descriptions from a LLM (\ie GPT-4), 
we conduct several ablation studies where we replace the LLM-generated region descriptions with (i) corresponding anatomical region names; (ii) random phrases of comparable length generated by LLM using the prompt: “Give me some random phrases related to [class] (around 50 words).”; (iii) very concise one-sentence summaries generated by LLM using the prompt: “[Ref Image] Please describe the [class] in a chest X-ray image in one single sentence.”; (iv) descriptions with subtle bias injected by LLM using the prompt: “[Description] Please introduce some minor errors into the anatomical description given above."; (v) descriptions written by board-certified clinicians. Then we re-train our pipeline using these modified descriptions.

The results in Table~\ref{tab:ablation} show that removing or weakening the medically meaningful content in the descriptions by using only anatomical region names (descriptions: LLM $\rightarrow$ names), random phrases of comparable length (description: LLM $\rightarrow$ random phrases), one-sentence summaries (description: LLM $\rightarrow$ one sentence), or subtly biased descriptions (description: with subtle bias) consistently leads to a significant performance drop compared with our default setting. This indicates that the anatomical knowledge provided by LLM is important and proves that the LLM-generated descriptions we use in practice provide sufficiently accurate and comprehensive guidance.
Moreover, using clinician-written descriptions (description: LLM $\rightarrow$ clinician) yields performance comparable to the LLM-generated ones, implying the effectiveness of the LLM-generated descriptions.

We also investigate the impact of using text descriptions generated by different LLMs. Specifically, we replace GPT-4 with other widely used LLMs, including Claude Sonnet 3.7 and DeepSeek-VL \cite{lu2024deepseek}. These variants are denoted as LLM: GPT $\rightarrow$ Claude and LLM: GPT $\rightarrow$ DeepSeek respectively in Table~\ref{tab:ablation}. We can observe that their performance is comparable to that of the default GPT-4 setting, demonstrating the robustness of our method. 

Moreover, we generate the descriptions by prompting the LLM with a reference image randomly sampled from the dataset and the text instruction. If we remove the reference image and only use the instruction (\ie LLM: text only in Table~\ref{tab:ablation}), we observe a slight performance degradation. If we instead generate descriptions for each image (\ie ref image: fixed $\rightarrow$ specific), we cannot obtain further performance improvements. It is important to note that the reference image is used only as a visual cue to better elicit the anatomical knowledge already stored in the LLM, rather than as a specific case to be diagnosed. In practice, we randomly select a representative reference image once and then keep it fixed; the resulting descriptions are region-specific but are not tied to any particular diseases. To further verify the robustness of our reference image selection strategy, we evaluate the impact of the reference image source. Specifically, we randomly sample three chest X-ray images from normal cases and three from abnormal cases, respectively. For each of these six reference images, we combine the image with the same instruction mentioned in Sec.~\ref{sec:feature_extractor} to prompt LLM and obtain region-wise anatomical descriptions. We then do three runs of our pipeline by using the three normal cases and three by using abnormal cases. As shown in Table~\ref{tab:ablation}, the averaged results demonstrate highly consistent performance between random (first row in Table~\ref{tab:ablation}), normal (ref image: normal in Table~\ref{tab:ablation}) and abnormal cases (ref image: abnormal in Table~\ref{tab:ablation}), supporting the robustness of our default strategy.



\noindent \textbf{Multimodal report generator.}
To validate the effect of the textual instruction in the multimodal report generator, we conduct an experiment by removing it.
This (Table~\ref{tab:ablation}: w/o $F_{inst}$) leads to a performance decrease on all metrics \eg, -0.028 in F1), demonstrating the necessity of the textual instruction in the multimodal report generator.

We further investigate the effectiveness of the visual transfer module in the multimodal report generator. As shown in Table~\ref{tab:ablation}, removing this module (\ie w/o VTrans)  
leads to a noticeable drop in performance (\eg, -0.018 in precision), highlighting its critical role in bridging the image encoder and the multimodal decoder. 
Without this module, the model struggles to effectively align the extracted visual features with the language modelling space, leading to a degradation in the quality and clinical relevance of the generated reports.


\noindent \textbf{Clinical classification and alignment.}
We further investigate the effectiveness of the proposed clinical classification and alignment scheme.

\noindent \emph{Clinical classification.}
We first remove the clinical classification head and its corresponding classification loss $\mathcal{L}_{disease}$ (see Table~\ref{tab:ablation}: w/o $\mathcal{L}_{disease}$). The results show that this leads to a clear performance decrease, \eg -0.058 in F1, indicating its benefits for improving the quality and clinical relevance of the generated reports.

\noindent \emph{Disease-oriented clinical alignment.}
We propose the disease-oriented clinical alignment scheme to align visual features not only with the textual features of the corresponding ground-truth report but also with those from other cases exhibiting similar diseases. 
In Table~\ref{tab:ablation} we use w/o DOCA as a variant without using this scheme in our framework. We can see that precision and F1 are substantially decreased from the original framework, \ie, -0.047 and -0.038, which proves the effectiveness of enhancing the MLLM with disease knowledge. 

\begin{figure}[t]
    \centering
    \includegraphics[width=\columnwidth]{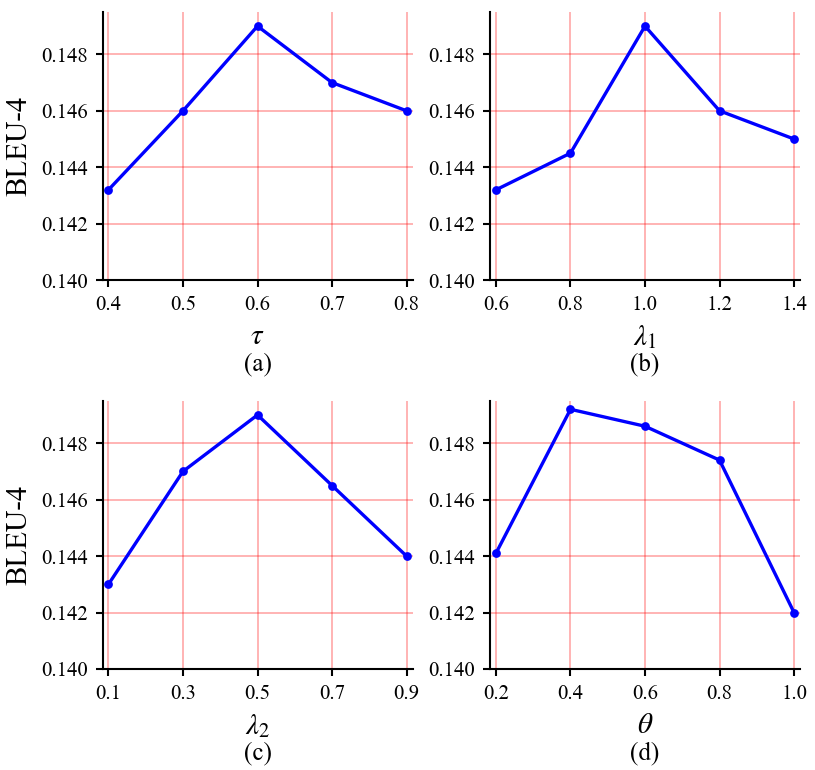}
    \caption{The effect of (a) similarity threshold $\tau$ used for selecting clinically relevant cases, (b) loss weight $\lambda_1$ for disease classification, (c) loss weight $\lambda_2$ for disease-oriented contrastive learning and (d) reward threshold $\theta$ for clinical quality reinforcement learning.}
    \label{fig:threshold_variation}
\end{figure}

Moreover, we retrieve similar cases based on disease labels in this scheme, if we follow previous works to select similar cases based on image-to-image or image-to-report similarity, as shown in Table~\ref{tab:ablation} (\ie DOCA $\rightarrow$ IIS and DOCA $\rightarrow$ IRS), there is a performance decrease on NLG and CE metrics. These results demonstrate that the proposed scheme can retrieve more clinically relevant cases, thereby helping the model better distinguishing different diseases and enhancing the clinical quality of the generated reports.

We also study the impact of disease correlation matrix on our model. As shown in Table~\ref{tab:ablation} (\mbox{\ie} DOCA: w/o disease matrix), when the disease correlation matrix is removed, there is a performance decrease on CE metrics. These results validate the effectiveness of modelling the correlations among diseases when performing disease-oriented clinical alignment.

In addition, we select similar cases within each training mini-batch. As shown in Table~\ref{tab:ablation}, extending the retrieval to the entire training dataset (\ie DOCA: batch $\rightarrow$ dataset) does not lead to further performance improvement. We speculate that the degradation mainly stems from the increased easy positive pairs introduced by global dataset retrieval. Specifically, DOCA relies on disease-level similarity to form positive/negative pairs. 
At the dataset scale, given a query case, its similar cases might be concentrated to a few highly homogeneous cases that are very similar in visual and textual patterns. 
In contrast, the positive pairs in batch sampling tend to be more diverse and challenging, so aligning them requires the model to learn and capture disease-relevant semantics. 
Therefore, global dataset retrieval can 
make DOCA focus on aligning easy cases, thereby reducing the difficulty of contrastive learning and limiting its ability to learn disease-relevant semantics.

\begin{figure*}[!t]
\includegraphics[width=\textwidth]{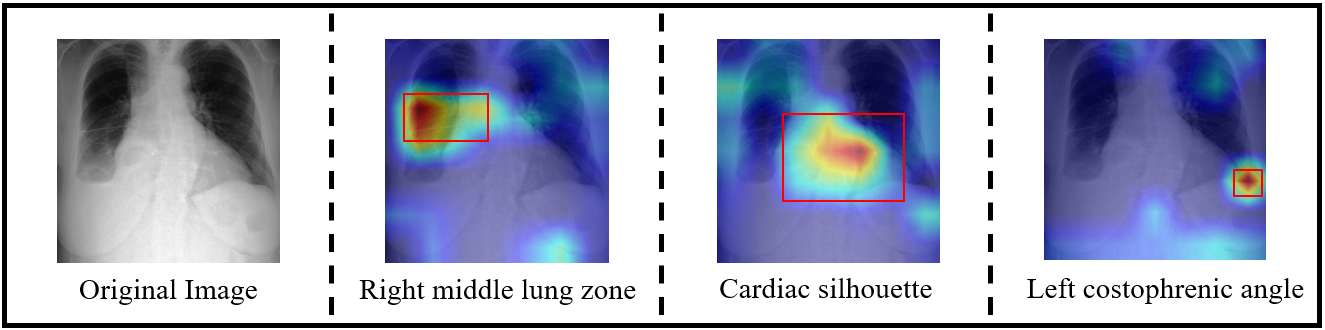}
\caption{Image-text attention heatmaps of our referring anatomical feature extractions. Colors from blue to red represent the attention weights from low to high. The red bounding boxes in heatmaps indicate the positions of the GT anatomical regions.}
\label{fig:referring_vis}
\end{figure*}

\begin{figure*}[htbp]
\includegraphics[width=\textwidth]{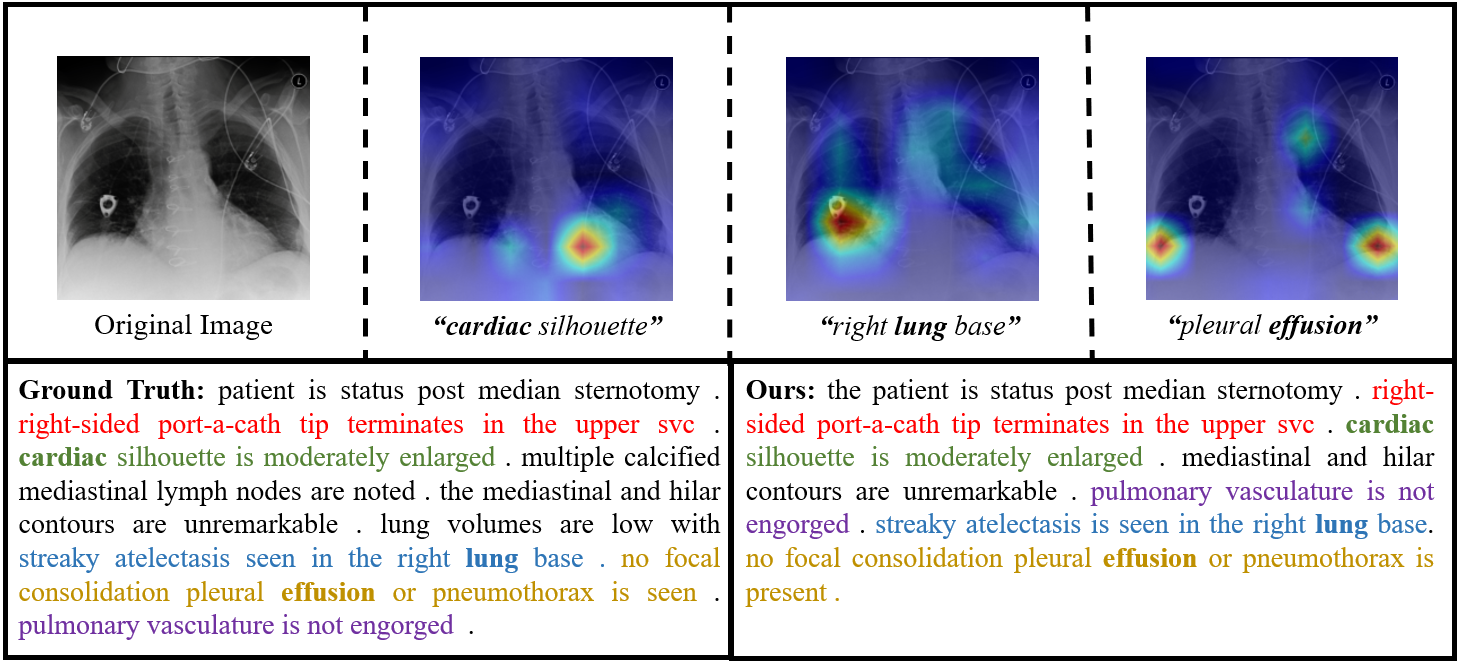}
\caption{An example of generated report with visualization of Grad-CAM on key medical terms (bold in text). Colors from blue to red represent the activation values from low to high, depending on which the model makes the prediction. Descriptions in the generated report that are consistent with the ground truth are highlighted in matching colors, and different anatomical regions are distinguished using different colors.}
\label{fig:report_vis}

\end{figure*}

In Figure~\ref{fig:threshold_variation} (a) we report the model performance under different similarity threshold $\tau$ used for selecting clinically relevant cases. The results show that the best performance is achieved when $\tau = 0.60$.

We vary the loss weights ($\lambda_1$ and $\lambda_2$) for the disease classification loss and the disease-oriented contrastive learning loss and report the performance in Figure~\ref{fig:threshold_variation} (b) and (c). We can observe that the best performance occurs when $\lambda_1=1$ and $\lambda_2=0.5$, which is our default setting.

\noindent \textbf{Clinical quality reinforcement learning.}
We further evaluate the advantage of our clinical quality reinforcement learning (CQRL) scheme. When we omit the CQRL process, in Table~\ref{tab:ablation} we find that the model's RadCliQ performance worsens significantly by 0.193, validating the effectiveness of enhancing the MLLM with report knowledge.

\noindent \emph{Reward function in RL.}
The reward function in the RL can be of other forms, \eg BLEU4.
In the Table~\ref{tab:ablation}, we find that the model's performance on BLEU4 can be improved by 0.005 when we replace RadCliQ with BLEU4, but the results on other metrics significantly decline, suggesting that the model is biased by specifically optimizing on BLEU4. In contrast, if we compare between w/o CQRL and MLLM-RRG in Table~\ref{tab:ablation}, by adding RL with RadCliQ, not only the result on RadCliQ is improved, the results on other metrics are also improved. The results suggest that our model can improve the clinical quality of generated reports during CQRL. 

\noindent \emph{Adaptive $\lambda_{KL}$.}
In the RL learning, if we change $\lambda_{KL}$ in Eqn.~\ref{equ8} to a fixed value of 1 (Table~\ref{tab:ablation}: w/ fixed $\lambda_{KL}$), it will result in a performance decrease, validating the usage of the adaptive $\lambda_{KL}$.

We also vary the controlling threshold $\theta$ in Eqn.~\ref{eq:9} for the adaptive KL weight and report the performance in Figure.~\ref{fig:threshold_variation} (d). We can observe that the best performance occurs when $\theta$ equals to 0.4, which is our default setting. 

\begin{figure*}[htbp]
\includegraphics[width=\textwidth]{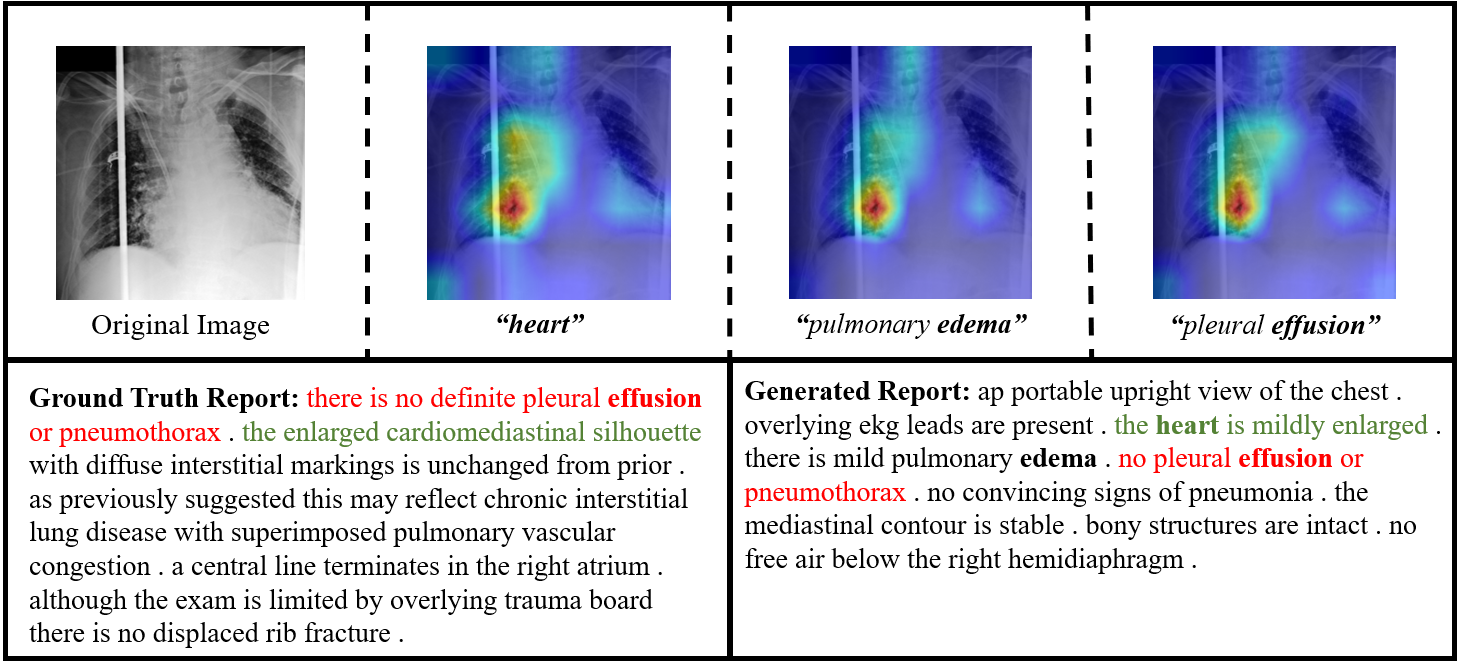}
\caption{An example of partially incorrect generated report due to the motion artifacts and occlusions in the chest X-ray image and visualizations of Grad-CAMs on key medical terms (bold in text). Colors from blue to red represent the activation values from low to high, depending on which the model makes the prediction. Descriptions in the generated reports that are consistent with the ground truth are highlighted in matching colors, and different anatomical regions are distinguished using different colors.}
\vspace{-0.5em}
\label{fig:incorrect_report_vis}
\end{figure*}

\begin{table*}[!t]
    \centering
    \caption{Effect of data cleaning on MIMIC-CXR and IU X-Ray for radiology report generation.}
    \begin{tabular}{l|c|ccc|ccc|c}
        \hline
        \multicolumn{1}{c|}{\multirow{2}{*}{Method}} & \multicolumn{1}{c|}{\multirow{2}{*}{Dataset}}  & \multicolumn{3}{c|}{NLG} & \multicolumn{3}{c|}{CE} & \multicolumn{1}{c}{\multirow{2}{*}{RadCliQ$\downarrow$}} \\
        \cline{3-8} 
        & & BLEU4$\uparrow$ & METEOR$\uparrow$ & ROUGE$\uparrow$ & Precision$\uparrow$ & Recall$\uparrow$ & F1$\uparrow$ & ~ \\ \hline
        MLLM-RRG & \multirow{2}{*}{MIMIC-CXR} & \textbf{0.131} & 0.174 & \textbf{0.299} & 0.529 & 0.550 & 0.539 & 0.825 \\
        \textbf{MLLM-RRG(Cleaned)} &  & 0.128 & \textbf{0.179} & \textbf{0.299} & \textbf{0.544} & \textbf{0.565} & \textbf{0.554} & \textbf{0.811} \\
        \hline
        MLLM-RRG & \multirow{2}{*}{IU X-Ray} & 0.235 & 0.230 & \textbf{0.417} & 0.752 & 0.687 & 0.718 & 0.501 \\
        \textbf{MLLM-RRG(Cleaned)} & & \textbf{0.240} & \textbf{0.237} & 0.413 & \textbf{0.768} & \textbf{0.708} & \textbf{0.737} & \textbf{0.460} \\
        \hline
    \end{tabular}
    \label{tab:data_cleaning}
\end{table*}

\subsection{Qualitative Analysis}
We further perform qualitative analysis on the MIMIC-CXR \mbox{\cite{johnson2019mimic}} dataset. First, we have visualized the last layer of the image-text attention in the referring anatomical feature extractor. As is shown in Figure~\ref{fig:referring_vis}, the regions of high attention weights basically overlap with ground truth anatomical regions, which proves that our model is clearly focused on the intended anatomical regions when extracting features. 
Second, we have created Grad-CAMs for our model on the test set of MIMIC-CXR. As shown in Figure~\ref{fig:report_vis}, the Grad-CAMs indicate our model consistently attends to the anatomically relevant regions when generating key medical terms. Meanwhile, our model can accurately localize abnormal regions and generate descriptions that closely align with the ground truth reports.

However, we acknowledge that it can be challenging for our model on low-quality chest X-ray images with substantial motion artifacts and occlusions (\mbox{\eg} overlying lines/tubes). As reflected in Figure~\ref{fig:incorrect_report_vis}, the model is distracted by the white vertical line (possibly the edge of the trauma board) and fails to generate accurate report or focus on the correct anatomical regions.

\section{Discussions}

\myparagraph{Data cleaning}
We observe that some reports in the original MIMIC-CXR and IU X-Ray datasets contain phrases that concern with past examination sequence (\mbox{\eg} “as compared to the previous radiograph” and "since prior study from"), which cannot be inferred from a single input image. Therefore, we first delete these phrases manually and then prompt an LLM (\mbox{\ie} Vicuna-7b\mbox{~\cite{chiang2023vicuna}}) to correct the grammar of the report: "You are a grammar correction assistant. Your task is to fix grammar errors ONLY. Correct grammar if needed: [Report]". Then we perform training/evaluation on the cleaned datasets.

Results in Table~\ref{tab:data_cleaning} show that after data cleaning, our model achieves improvements on most metrics for both MIMIC-CXR and IU X-Ray datasets in RRG task, especially on CE metrics and RadCliQ (\eg +0.015 in F1 on MIMIC-CXR and +0.019 in F1 on IU X-Ray). This validates the effect of our data cleaning procedure and proves that our model is not overfitted to irrelevant phrases for higher scores.

However, to ensure a fair comparison with state of the art methods, in Sec.~\ref{sec::model_comparison} we keep the original datasets and official splits as our default setting, which is consistent with prior works\mbox{~\cite{chen2020generating, xiang2024gmod, yi2024lhr}}

\myparagraph{Disease label re-annotation}
In our study, we follow the common practice~\cite{jin2024promptmrg,park2025dart} to obtain disease labels for MIMIC-CXR~\cite{johnson2019mimic} by directly applying CheXbert~\cite{smit2020chexbert} to each case and extracting the corresponding disease labels. Since CheXbert is primarily trained on MIMIC-CXR, this process is well aligned with the MIMIC-CXR data. However, directly applying CheXbert to IU X-Ray~\cite{demner2016preparing} may introduce a domain gap. To mitigate this issue, we first invite an on-board clinician to re-annotate 400 cases from IU X-Ray. Then we finetune CheXbert on this subset by updating only its final linear layer. Finally, we use the adapted model, namely IU-CheXbert, to extract disease labels for the remaining cases of IU X-Ray.

To validate the effectiveness of IU-Chexbert, we randomly split our re-annotated IU X-Ray cases into 80\% for training and 20\% for testing. Table~\ref{tab:iubert} reports the disease classification results on the test set of our re-annotated cases, where the adapted model (IU-CheXbert) improves F1 by +0.055 over the original CheXbert. The results prove that finetuning CheXbert on our re-annotated cases can effectively reduce the domain gap between IU X-Ray and MIMIC-CXR datasets.

\begin{table}[!t]\small
    \centering
    \caption{Disease classification results of CheXbert and IU-CheXbert on the test set of our re-annotated IU X-Ray cases.}
    \begin{tabular}{l|ccc}
       \hline
       Method & Precision$\uparrow$ & Recall$\uparrow$ & F1$\uparrow$ \\
       \hline
       CheXbert & 0.823 & 0.863 & 0.843 \\
       \textbf{IU-CheXbert(Ours)} & \textbf{0.914} & \textbf{0.883} & \textbf{0.898} \\
       \hline
    \end{tabular}
    \label{tab:iubert}
\end{table}

\myparagraph{Clinical study}
We have also invited a board-certified clinician to assess the quality of generated reports. Specifically, we randomly sample 200 reports from MIMIC-CXR test set. For each report, the clinician carefully reviews our generated report and scores each generated report on a 1-10 scale to access the accuracy of key findings and impressions: minor wording differences that do not affect patient management incur little or no penalty, whereas errors or omissions that could possibly change downstream diagnosis or treatment are heavily penalized. Statistical analysis shows that the average score for generated reports is 8.615$\pm$0.195, demonstrating their reliability and clinical validity.

\section{Conclusion}
In this work, we present a novel approach MLLM-RRG that combines MLLMs with clinical knowledge enhancement to generate sensible and accurate chest X-ray reports. 
Our method consists of a referring anatomical feature extractor captures visual features from both specific anatomical regions and the entire image; a multimodal report generator that comprises the visual transfer module and the multimodal decoder is developed to produce coherent report; Additionally, we perform disease-oriented clinical classification and alignment in a multi-task learning framework to improve the clinical significance of the generated report; 
we also employ a reinforcement learning mechanism with RadCliQ as the reward function to enhance the clinical accuracy of the generated reports. 
In above ways, we enhance the MLLM with anatomical, disease and report knowledge, enabling it to generate clinically accurate and professionally standardized radiology reports.
Results have shown that our approach surpasses state of the art on the MIMIC-CXR and IU X-Ray datasets.
In the future, we plan to investigate how variations in imaging devices and acquisition protocols may affect model performance, and to systematically evaluate and improve the model’s generalization across different patient populations and ethnic groups.

\section*{REFERENCES}
\vspace{-1.5em}

\bibliographystyle{IEEEtran}
\bibliography{cas-refs}

\end{document}